\documentclass[conference]{IEEEtran}
\IEEEoverridecommandlockouts
\usepackage{cite}
\usepackage{amsmath,amssymb,amsfonts}
\usepackage[ruled,vlined]{algorithm2e}
\usepackage{graphicx}
\usepackage{textcomp}
\usepackage{placeins}
\usepackage{xcolor}
\usepackage{url}
\usepackage{subcaption}
\usepackage{array}
\usepackage{booktabs}
\usepackage{multirow}
\usepackage[colorlinks=true, linkcolor=blue, urlcolor=blue]{hyperref}
\def\BibTeX{{\rm B\kern-.05em{\sc i\kern-.025em b}\kern-.08em
    T\kern-.1667em\lower.7ex\hbox{E}\kern-.125emX}}

\title{\vspace{0.5in} \LARGE \bf Development of an Autonomous Mobile Robotic System for Efficient and Precise Disinfection}
\author{Ting-Wei Ou$^{1}$, Jia-Hao Jiang$^{2}$, Guan-Lin Huang$^{3}$ and Kuu-Young Young$^{4}$, Senior Member, IEEE
\thanks{$^{1}$Ting-Wei Ou, Graduate Degree Program of Robotics, National Yang Ming Chiao Tung University, Hsinchu, Taiwan.
{\tt\small kklb1716@gmail.com}}%
\thanks{$^{2}$Jia-Hao Jiang, Institute of Electrical and Control Engineering, National Yang Ming Chiao Tung University, Hsinchu, Taiwan.
    {\tt\small dino95002@gmail.com}}%
\thanks{$^{3}$Guan-Lin Huang, Department of Mechanical Engineering, National Yang Ming Chiao Tung University, Hsinchu, Taiwan.
{\tt\small huang20040926@gmail.com}}%
\thanks{$^{4}$Kuu-Young Young(corresponding author), professor with Department of Electrical Engineering, National Yang Ming Chiao Tung University, Hsinchu, Taiwan.
    {\tt\small kyoung@nycu.edu.tw}}%
}

\begin{document}
\maketitle
\begin{abstract}
The COVID-19 pandemic has severely affected public health, healthcare systems, and daily life, especially amid resource shortages and limited workers. This crisis has underscored the urgent need for automation in hospital environments, particularly disinfection, which is crucial to controlling virus transmission and improving the safety of healthcare personnel and patients. Ultraviolet (UV) light disinfection, known for its high efficiency, has been widely adopted in hospital settings. However, most existing research focuses on maximizing UV coverage while paying little attention to the impact of human activity on virus distribution. To address this issue, we propose a mobile robotic system for UV disinfection focusing on the virus hotspot. The system prioritizes disinfection in high-risk areas and employs an approach for optimized UV dosage to ensure that all surfaces receive an adequate level of UV exposure while significantly reducing disinfection time. It not only improves disinfection efficiency but also minimizes unnecessary exposure in low-risk areas. In two representative hospital scenarios, our method achieves the same disinfection effectiveness while reducing disinfection time by 30.7\% and 31.9\%, respectively. The video of the experiment is available at: \href{https://youtu.be/wHcWzOcoMPM}{https://youtu.be/wHcWzOcoMPM}.

\end{abstract}

\begin{IEEEkeywords}
 UV Disinfection, Mobile Robot Manipulator, Hotspot Identification, UV Dosage Optimization 
\end{IEEEkeywords}
%%%%%%%%%%%%%%%%%%%%%%%%%%%%%%%%%%%%%%%%%%%%%%%%%%%%%%%%%%%%%%%%%%%%%%%%%%%%%%%%0
\section{Introduction}

\begin{figure}[t] 
    \centering
    \includegraphics[width=0.48\textwidth]{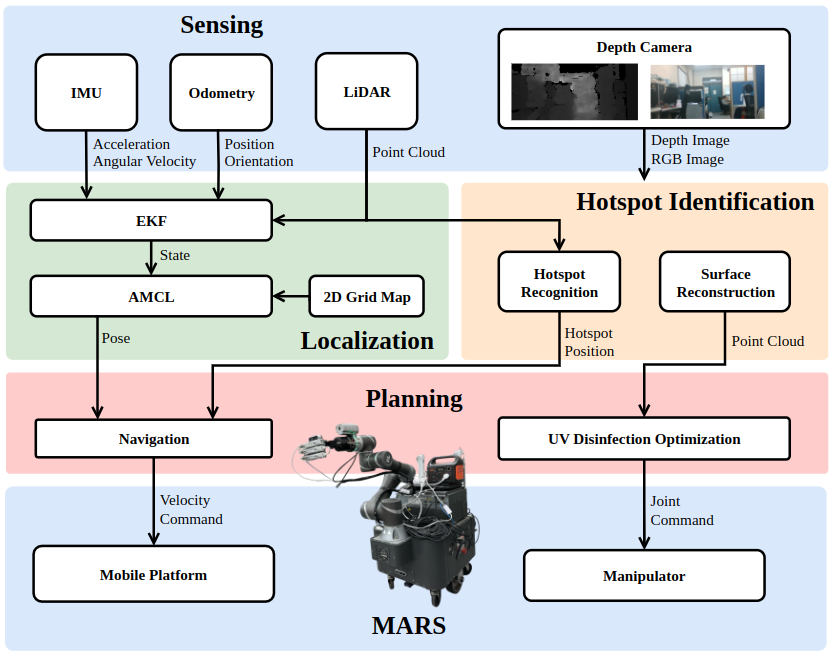} 
    \caption{System architecture of the mobile robotic UV disinfection system.} 
    \label{fig:Fig1} 
\end{figure}

The shortage of human resources has long been an urgent problem in medical settings and has been further exacerbated by the COVID-19 pandemic, which has significantly increased the workload of healthcare personnel, including disinfection. Traditional manual disinfection methods are not only labor-intensive and time-consuming but also have a high risk of cross-contamination. To improve both efficiency and safety during operation, UV-C (Ultraviolet-C) disinfection robots \cite{b1, b2, b3} have gained widespread attention as an automated and sustainable means. Meanwhile, ensuring that these robots can navigate complex hospital environments in a flexible way and perform disinfection tasks efficiently and accurately remains a critical challenge \cite{b4}.

Several studies used optimal path planning and deployment strategies \cite{b5,b6} to maximize disinfection effectiveness. However, dynamic changes in medical environments strongly influence the efficiency of robotic disinfection. As a key factor, human activity causes fluctuations in viral concentrations, leading to varying demands in different areas \cite{b7,b8}. As a result, integrating environmental perception into disinfection strategies to adapt to different risk zones has attracted a great deal of attention. Some research applied semantic recognition to identify viral hotspots of high risk \cite{b9,b10,b11}, allowing robots to perform precise and targeted disinfection. However, most of them focus mainly on global path planning and optimizing disinfection coverage, but not on dynamically adjusting UV dosage according to environmental risk levels. In particular, high-contact surfaces (e.g., door handles, bed rails) pose a significantly higher risk of viral contamination than others (e.g., curtains, walls), necessitating a more adaptive and risk-based disinfection strategy. 

Based on the system previously developed in our laboratory \cite{b12}, which employed a disinfection approach without differentiating surface-specific risks, this research proposes a strategy that can dynamically adjust the corresponding UV exposure according to the degree of contamination. To achieve this, we start with the identification of disinfection hotspots and distinguishing high-contact surfaces from others for risk-aware planning. Then a strategy for optimizing UV dosage that corresponds to varying risk levels is developed that ensures sufficient disinfection while optimizing efficiency. Our real-world disinfection experiments show that the proposed system reduced disinfection time by approximately 30\% compared to non-hotspot-based approaches while achieving the disinfection requirement across all surfaces.

The remainder of the paper is organized as follows: the proposed disinfection system is described in Section 2, and the hotspot disinfection strategy is outlined in Section 3. The experimental results are given in Section 4. Finally, the concluding remarks are presented in Section 5.

%%%%%%%%%%%%%%%%%%%%%%%%%%%%%%%%%%%%%%%%%%%%%%%%%%%%%%%%%%%%%%%%%%%%%%%%%%%%%%%%
\section{Proposed Disinfection System}
To enable robots to perform disinfection tasks and to tackle the uneven distribution of viruses caused by human activity efficiently \cite{b7}, \cite{b8}, an autonomous disinfection system is proposed, consisting of four main modules: Sensing, Hotspot Identification, Localization, and Planning, as illustrated in Figure \ref{fig:Fig1}. The sensing module provides information about the environment and the robot. The hotspot identification module searches for high-risk areas and then constructs a corresponding 3D representation of the environment through two key functions: hotspot recognition and surface reconstruction. The first employs YOLOv8 \cite{b13}, which outperforms previous versions in precision and real-time detection, to identify potential contamination areas and determine optimal stopping points for path planning. The latter integrates semantically segmented depth images with the manipulator's end effector pose and incorporates them into an OctoMap \cite{b17}, a probabilistic 3D mapping framework based on octrees, which yields efficient volumetric representation, allowing the system to maintain a compact yet detailed map of the environment.

The localization module is used to ensure accurate robot movement using the adaptive Monte Carlo localization algorithm (AMCL) \cite{b20}, which compares real-time sensor data with a preconstructed LiDAR map for precise localization. It incorporates an Extended Kalman Filter (EKF) to improve stability by fusing wheel odometry and IMU data. Finally, the planning module deals with robot chassis navigation and UV disinfection optimization. To efficiently reach each designated location, the most suitable sequence of hotspots is determined on the basis of their spatial distribution. And, for chassis navigation, we employ the A* algorithm \cite{b21} combined with the Dynamic Window Approach (DWA) \cite{b22} for path planning. For manipulator motion planning, we propose a hotspot disinfection optimizer, discussed in the following section, to calculate the optimal ultraviolet dose and the manipulator's motion speed, so that the process can be completed effectively while meeting the disinfection demand. 

%%%%%%%%%%%%%%%%%%%%%%%%%%%%%%%%%%%%%%%%%%%%%%%%%%%%%%%%%%%%%%%%%%%%%%%%%%%%%%%%
\section{Hotspot Disinfection Strategy}
The proposed strategy is intended to improve overall disinfection efficiency by flexibly distributing UV dose between hot and non-hot zones. It includes three main functions: hotspot recognition, surface preprocessing, and UV disinfection optimization. That of hotspot navigation determines the traversal paths for hotspot zones, surface preprocessing reconstructs the hotspot object's surface for manipulator's path planning, and UV disinfection optimization adjusts the speed of the manipulator to achieve effective dosage while maximizing efficiency. Appropriate disinfection standards are established to ensure sufficient UV exposure for different surfaces. Based on research on UV dosage and COVID-19 inactivation \cite{b15}, we set the dosage standards for hotspot and non-hotspot surfaces as 22 mJ/cm² (for intensive virus inactivation) and 5 mJ/cm² (for basic virus inactivation), respectively.

\subsection{Hotspot Recognition}
\begin{figure}[t]
    \centering
    \begin{subfigure}[b]{0.155\textwidth} % 1/3 of 0.5\textwidth
        \centering
        \includegraphics[width=\textwidth]{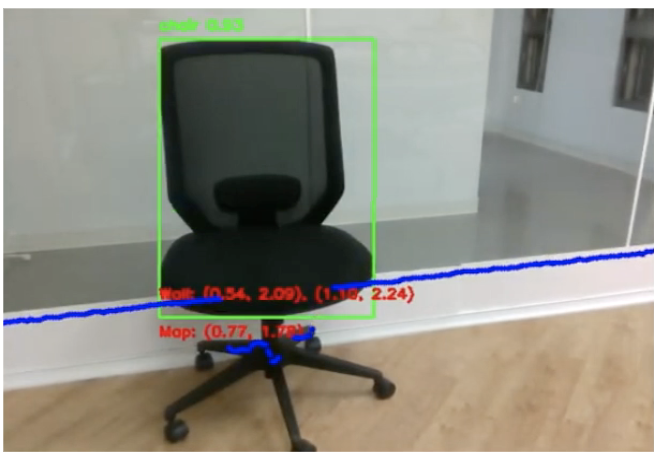}
        \caption{}
    \end{subfigure}
    \begin{subfigure}[b]{0.155\textwidth}
        \centering
        \includegraphics[width=\textwidth]{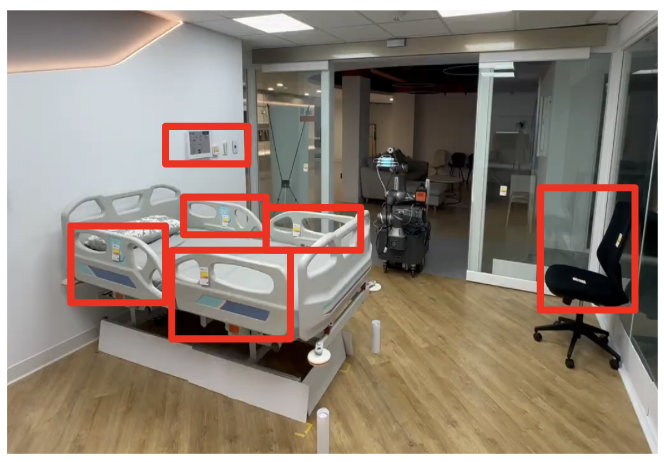}
        \caption{}
    \end{subfigure}
    \begin{subfigure}[b]{0.155\textwidth}
        \centering
        \includegraphics[width=\textwidth]{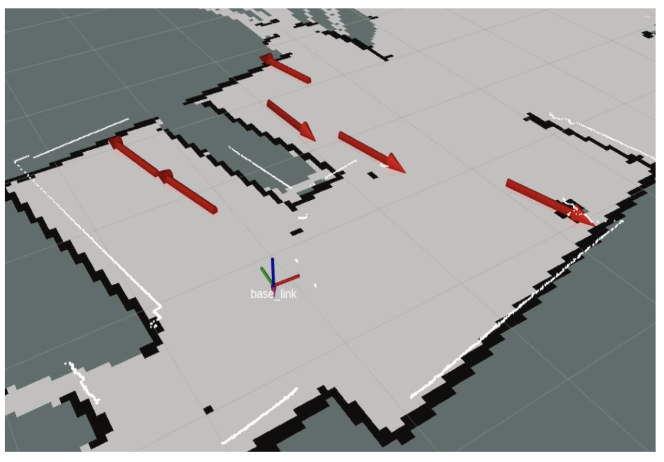}
        \caption{}
    \end{subfigure}
    \caption{(a) Projecting LiDAR points (blue) onto the camera image to obtain the depth of the detected hotspot (green box). (b) Identified hotspot areas in a patient ward (red box). (c) Mapping the detected hotspot areas onto the grid map.}
    \label{fig:Fig2}
\end{figure}

The distribution of viruses is positively correlated with the likelihood of surface contact \cite{b7}, \cite{b8}. Consequently, armless office chairs, examination tables, side rails, medical machines, switches, sinks, tables, etc., which are considered frequently touched objects, require a higher UV dose for disinfection. For their precise identification, the ADE20K dataset\cite{b14}, which includes a wide range of indoor and outdoor objects, is utilized for training, along with 1,600 manually annotated object samples related to medical environments collected by our laboratory to supplement the lack of medical-specific objects in the original ADE20K dataset. With hotspot information available for disinfection, the robot will first scan the environment to identify high-risk objects and then mark them on a grid map, as shown in Figure \ref{fig:Fig2}. 

\begin{figure}[t] 
    \centering
    \includegraphics[width=0.45\textwidth]{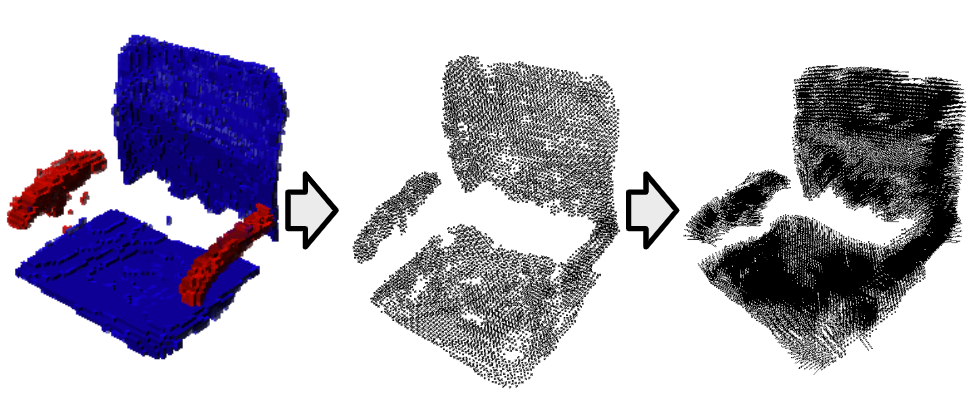} 
    \caption{Preprocessing of high-touch surfaces: constructing the OctoMap, sampling and filtering the point cloud, and then estimating the surface normals.} 
    \label{fig:Fig3} 
\end{figure}

\begin{figure}[h] 
    \centering
    \includegraphics[width=0.4\textwidth]{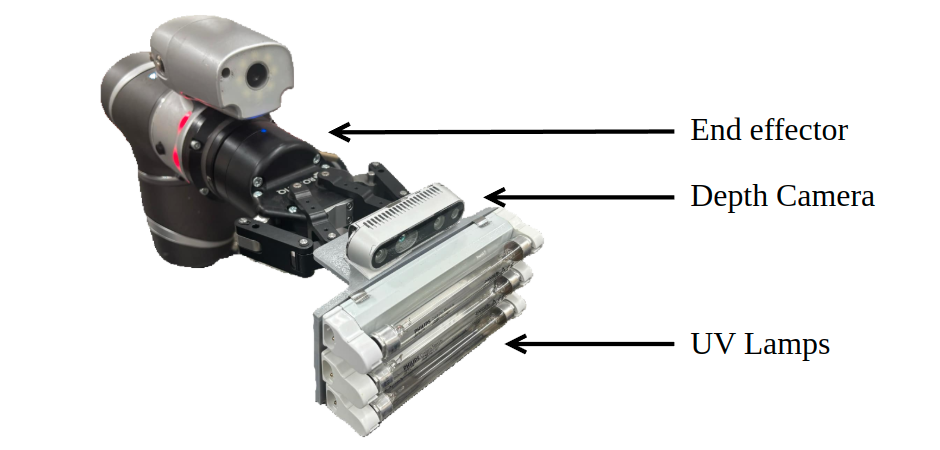} 
    \caption{Three UV-C lamps attached to the end effector for intensive radiation.} 
    \label{fig:Fig4} 
\end{figure}

\begin{figure}[h] 
    \centering
    \includegraphics[width=0.46\textwidth]{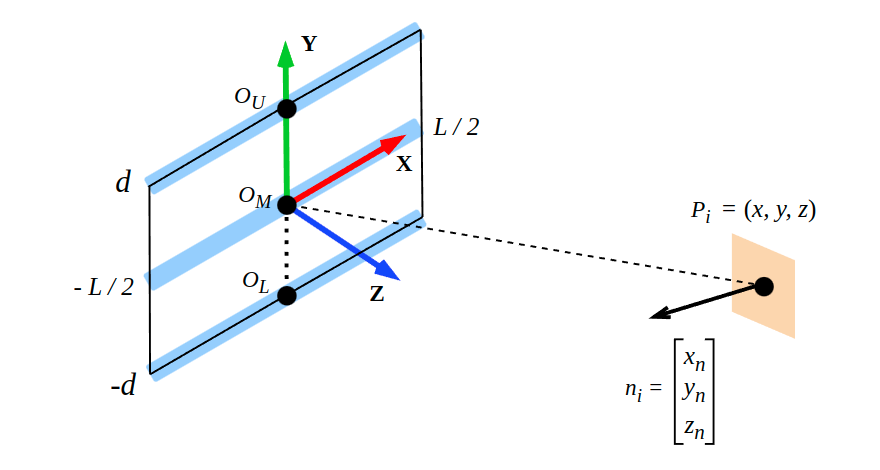} 
    \caption{Geometric relationship between UV-C lamps and the target surface.} 
    \label{fig:Fig5} 
\end{figure}

\subsection{Surface Preprocessing}
The system uses OctoMap \cite{b17} and the pose of the end effector to reconstruct the surface of the object, resulting in an accurate 3D point cloud, which is used for path planning and UV dosage estimation, as illustrated in Figure \ref{fig:Fig3}. To prevent the manipulator from colliding with objects, the path is designed to maintain a constant distance of 30 cm from the surface of the object. In addition, points on the path are chosen from a precomputed set of reachable positions to avoid singularity. 

\subsection{UV Disinfection Optimization}
 To ensure adequate radiation coverage for all surface points while minimizing total disinfection time, we propose a strategy to optimize the speed of the manipulator in following the planned path.  It first estimates the dose of the UV disinfection device on the target surface. Then a radiation model based on \cite{b16} is developed to evaluate the irradiance. The device attached to the end effector is formed by three connected UV lamps, as shown in Figure \ref{fig:Fig4}, with the geometric relationship to the surface illustrated in Figure \ref{fig:Fig5}, where the centers of the upper, middle, and lower lamps are located at \( O_U (0, d, 0) \), \( O_M (0, 0, 0) \), and \( O_L (0, -d, 0) \), respectively, and \( P_i = (x, y, z) \) is a point on the surface. Assume that the total luminous flux of a lamp, denoted by \( \Phi_E \), is modeled as a uniform linear light source of length \( L \). With the lamp divided into small segments of length \( \Delta l \), it yields a luminous flux of \( \Phi_E/L \cdot \Delta l \) for each segment. As \( \Delta l \) approaches zero, these segments can be treated as point light sources. The illuminance at a given point on the surface can then be formulated as an integral throughout the length of the lamp, expressed in Equation (1):

\begin{equation}
\mathrm{d}E = \frac{1}{4\pi r'^2} \frac{\Phi_E}{L} \mathrm{d}\ell \
= \frac{\Phi_E}{4\pi L} \frac{1}{(x - \ell )^2 + (y-y')^2 + z^2} \mathrm{d}\ell \
\end{equation}

\noindent where \( r' \) denotes the distance between the light source and the target point, \( \ell \) the position of a light source on the lamp, and \( y' \) the installation position of the lamp along the \( y \)-axis. Considering further the direction of radiation relative to the normal of the surface, \( \mathbf{\hat{r}} \cdot \mathbf{\hat{n}} \) 
(\( [x - \ell, y - y', z] [x_n, y_n, z_n]^\top \)) is included, and Equation (1) is rewritten as

\begin{align}
\mathrm{d}E = \frac{\Phi_E}{4\pi L} \frac{1}{(x - \ell)^2 + (y-y')^2 + z^2} \mathrm{d}\ell \, \hat{r} \cdot \hat{n}
\end{align}

\noindent By integrating from \( -\ell/2 \) to \( \ell/2 \), the entire illuminance at a surface point is calculated as 

\begin{equation}
\begin{aligned}
E &= \frac{\Phi_E}{4 \pi L \sqrt{(y-y')^2 + z^2}} 
\bigg(
\arctan\left(\frac{\frac{L}{2} - x}{\sqrt{(y-y')^2 + z^2}}\right) \\
&\qquad + \arctan\left(\frac{\frac{L}{2} + x}{\sqrt{(y-y')^2 + z^2}}\right)
\bigg) \hat{r} \cdot \hat{n}
\end{aligned}
\end{equation}

\noindent Consequently, the total illuminance $E_{\text{total}}$ for a surface point from the three lamps can be derived accordingly. 

With available dose estimation, we went on to develop Algorithm~\ref{alg:example} to optimize movement speed along the planned path under the requirement that each surface point \( P_n \) must receive at least minimum dosage for both hotspot and non-hotspot surfaces. We formulated this optimization problem as a linear programming one and adopted the interior-point method \cite{b18} to solve it, which is known for its numerical stability and robust performance. With Algorithm~\ref{alg:example}, the system can quickly determine the optimal speed profile along the path, ensuring that each target surface point receives the required UV dose. 

\begin{algorithm}
\caption{Manipulator Speed Optimization}\label{alg:example}
\KwIn{$\mathcal{J}$: Set of precomputed motion trajectories \\
\Indp \Indp $D_{\min}$: Minimum UV dose \\
$\mathcal{P}$: Set of surface points}
\KwOut{$\mathcal{U}$: Set of optimized velocities}

Initialize each time allocation $T = \{\Delta t_i \geq 0.1\}$ \\
Set objective = Minimize total exposure time

\For{each surface point $P_n \in \mathcal{P}$}{
    total\_dosage $\gets 0$ \\
    \For{each trajectory $J_n \in \mathcal{J}$}{
        $E_{\text{total}} \gets \text{dosage\_estimate}(J_n, P_n)$ \\
        total\_dosage $\gets$ total\_dosage + $E_{\text{total}} \cdot T_n$
    }
    Add constraint: $D_n \geq D_{\min}$
}
Optimize $(T)$ \\
$\mathcal{U}$ $\gets$ getSpeed($\mathcal{J}, T$) \\
\Return $\mathcal{U}$

\end{algorithm}

%%%%%%%%%%%%%%%%%%%%%%%%%%%%%%%%%%%%%%%%%%%%%%%%%%%%%%%%%%%%%%%%%%%%%%%%%%%%%%%%
\begin{figure*}[t] 
    \centering
    \includegraphics[width=0.95\textwidth]{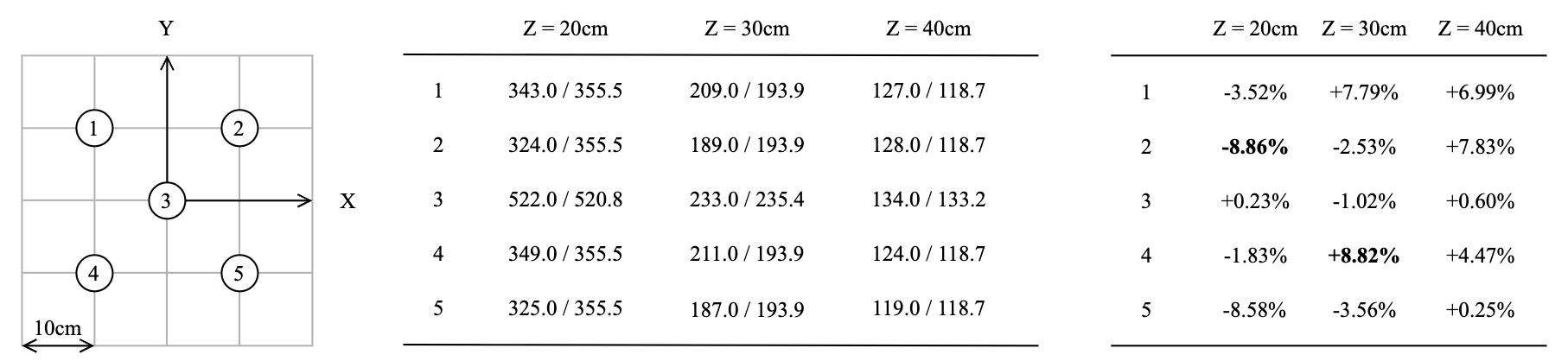} 
    \caption{(Left) Sensor placement positions on the XY-plane. (Middle) Measured vs. theoretical irradiance values ($\mu\text{W}/\text{cm}^2$) at different Z distances. (Right) Relative error (\%) at each position. All deviations are within 10\%, which is within the specified tolerance of the UV-C sensor.} 
    \label{fig:Fig6} 
\end{figure*}

\begin{figure*}[t] 
    \centering
    \includegraphics[width=0.95\textwidth]{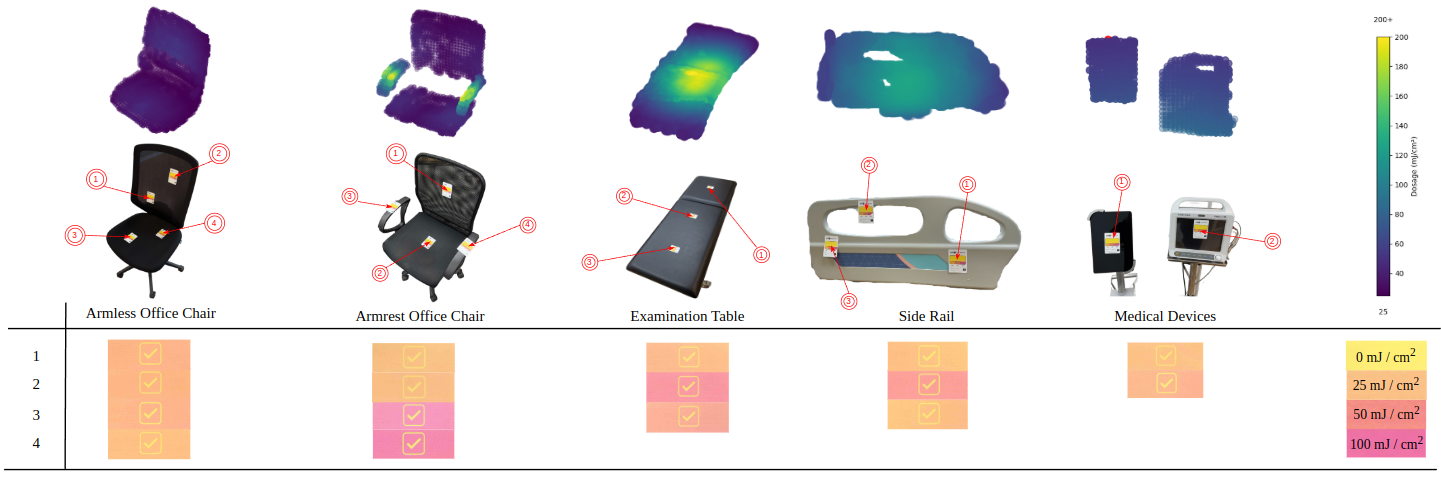} 
    \caption{Results for dose estimation: (Upper) The estimated hotspot dose distribution by the proposed dose estimation method; (Middle) The placement of hotspot objects and dosimeter cards; (Bottom) The color change of the dosimeter cards after disinfection. The results show that the color change trend of the dosimeter card aligns with the estimation.} 
    \label{fig:Fig7} 
\end{figure*}

\newpage
\section{Experiments}
We conducted a series of experiments to evaluate the performance of the proposed autonomous disinfection system. The mobile robot manipulator (MARS), developed by the Industrial Technology Research Institute, Taiwan, was adopted as the disinfection robot, as shown in Figure \ref{fig:Fig1}. It consists of a differential drive chassis (60 cm × 80 cm × 90 cm), a Techman TM5-700 robotic arm, and several sensors. For localization, the system used a Hokuyo UTM-30LX-EW 2D LiDAR and an Intel D435 depth camera. The computing unit, powered by an Intel i5-12500H processor, an RTX 2050 GPU, and 16 GB of RAM, ensures real-time sensor processing and motion planning. Three Philips 4W UV-C lamps were mounted on the end effector of the robot, as shown in Figure \ref{fig:Fig4}, which were especially intended for hotspot disinfection, along with two Philips 15W UV-C lamps on the back to assist disinfection in general areas. And, UV-C 254 TRI Card Dosimeters \cite{b23} are deployed to measure the dose received by the surfaces.

\begin{figure}[t]
    \centering
    \begin{subfigure}{0.16\textwidth}
        \centering
        \includegraphics[width=\textwidth]{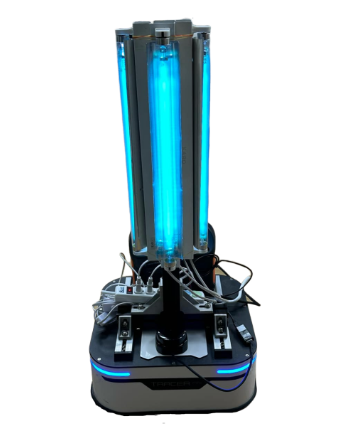}
        \label{fig:image1}
    \end{subfigure}
    \caption{Towerbot.}
    \label{fig:Fig8}
\end{figure}

\begin{figure*}[t] 
    \centering
    \includegraphics[width=0.88\textwidth]{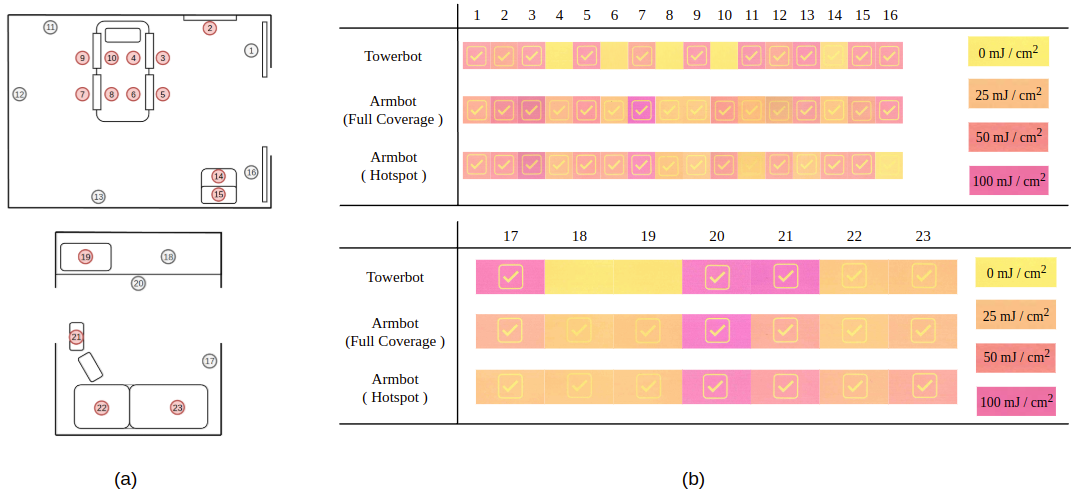} 
    \caption{(a) Placement and (b) measurement results of dosimeter cards. The upper image is a single-patient ward; the bottom image is a clinic room. Red numbers: hotspot objects.} 
    \label{fig:Fig9} 
\end{figure*}

The first two sets of experiments were conducted to evaluate whether the proposed dose estimation method can accurately approximate actual UV exposure. First, we used an RGM-UVC LCD Digital Lamp Light Meter to validate the UV-C irradiance predicted by our irradiation model in Equation (3). The sensor was placed at five different positions on the XY-plane, and measurements were taken with the end-effector positioned along the Z-axis at distances of 20, 30, and 40 cm from the plane. The results, shown in Figure \ref{fig:Fig6}, indicate that the maximum deviation between the measured and theoretical values was within 10\%, which falls within the measurement tolerance of the UV-C sensor. This confirms the accuracy of the proposed irradiation model.
 
 Second, to evaluate the actual dosage on different objects' surfaces, we placed dosimeter cards on five objects, including an armless office chair, an examination table, a side rail, and two medical devices. The system then performed surface reconstruction to generate the surface of each object for disinfection. Subsequently, the UV disinfection optimization method was applied to plan the path considering both speed and dosage. After execution, the color change of the dosimeter cards was analyzed to derive the actual UV exposure in each area and compared to the estimated one using our proposed method. As shown in Figure \ref{fig:Fig7}, the upper and middle parts illustrate the estimated dose distribution and placement of the dosimeter cards, where different colors in the point clouds represent the estimated UV doses, and the bottom part shows the color change on the dosimeter cards. Because the maximum detectable dose of the cards was 100 mJ/cm², regions with a dose exceeding this value could not be demonstrated by a color change, while it did not affect the analysis. From the results, it was shown that the proposed method well predicted the distribution of UV-C doses and ensured that the minimum dose in all marked regions reached 25 mJ/cm², the required threshold for viral inactivation. 

The second set of experiments was designed to evaluate the effectiveness of the proposed disinfection system in a medical environment. It was carried out in the NYCU intelligent ward, which was specifically designed to test intelligent medical systems developed by research teams at our university. To provide a comparison, two other robotic disinfection systems were also applied for the experiments. The first, named Towerbot and shown in Figure \ref{fig:Fig8}, was currently the most popular, adopting a strategy that followed a fixed path with a fixed time during disinfection. Because it was not attached to an arm, the surfaces of the objects on the back side were expected to receive less illuminance. The second, named Ambot-F, was previously developed in our laboratory \cite{b12},  which was intended to achieve thorough and even disinfection of the surfaces of all objects in a 3D manner. Both Ambot-F and the proposed system (named Ambot-H) adopted the same mobile robot manipulator, the MARS.  Towerbot was equipped with four 15W UV-C lamps to be compatible with those on MARS.

We selected two representative hospital scenarios for testing: a single-patient ward and a clinic room. UV dosimeter cards were evenly distributed in both rooms, and the environments were divided into hotspot and non-hotspot areas, as shown in Figure \ref{fig:Fig9}(a). In Figure \ref{fig:Fig9}(b), the Towerbot results show that the color changes on cards 4, 6, 8, 10, 18, and 19 were not evident. This might be due to an incomplete coverage of the surfaces of the objects, even with enough exposure time. In comparison, both Armbot-F and Armbot-H achieved a similar overall disinfection effect. Here, cards 11 and 16 for Armbot-H did not reach 25 mJ/cm² for being placed in non-high-risk areas, which consequently demonstrated its ability to distinguish the environment for disinfection. 

For further analysis, we investigated three indices: hotspot disinfection coverage rate (HCR, exceeding 25 mJ/cm²), overall disinfection coverage rate (OCR, exceeding 5 mJ/cm²), and execution time (ET). The results for the patient ward and the clinic room are listed in Tables~\ref{tab:1} and \ref{tab:2}, respectively. In Table~\ref{tab:1}, Towerbot achieved HCR and OCR of 54.5\% and 75\%, respectively, with an ET of 55 min. This indicates that Towerbot was prone to blind spots in environments with obstacles, causing some areas to receive insufficient UV exposure. Both Armbot-F and Armbot-H achieved 100\% in HCR and OCR. However, Armbot-H reduced the execution time to 24.26 min compared to 35.15 min for Armbot-F, saving 30.7\% of the time by reducing time spent in non-hotspot areas.
In Table~\ref{tab:2}, for the clinic room, Towerbot achieved HCR and OCR of 75\% and 71.4\%, respectively, with an ET of 30 min. Again, both Armbot-F and Armbot-H reached 100\% in HCR and OCR. Armbot-F completed disinfection in 20.49 min, while Armbot-H further reduced the time to 14.10 min, saving 31.9\% of the time.
These results demonstrate that the proposed Armbot-H successfully achieved efficient and precise disinfection by properly distributing UV dosage across high- and low-risk areas.

\begin{table}[t]
    \centering
    \renewcommand{\arraystretch}{1.3}
    \caption{Results in the patient ward.}
    \label{tab:1}
    \begin{tabular}{lccc}
        \textbf{} & \textbf{HCR} & \textbf{OCR} & \textbf{ET} \\
        \midrule
        Towerbot & 54.5\% & 75\% & 55:00 \\
        Armbot-F & \textbf{100\%} & \textbf{100\%} & 35:15 \\
        Armbot-H & \textbf{100\%} & \textbf{100\%} & \textbf{24:26} \\
        \midrule
    \end{tabular}
\end{table}

\begin{table}[t]
    \centering
    \renewcommand{\arraystretch}{1.3}
    \caption{Results in the clinic room.}
    \label{tab:2}
    \begin{tabular}{lccc}
        \textbf{} & \textbf{HCR} & \textbf{OCR} & \textbf{ET} \\
        \midrule
        Towerbot & 75\% & 71.4\% & 30:00 \\
        Armbot-F & \textbf{100\%} & \textbf{100\%} & 20:49 \\
        Armbot-H & \textbf{100\%} & \textbf{100\%} & \textbf{14:10} \\
        \midrule
    \end{tabular}
\end{table}

%%%%%%%%%%%%%%%%%%%%%%%%%%%%%%%%%%%%%%%%%%%%%%%%%%%%%%%%%%%%%%%%%%%%%%%%%%%%%%%%
\section{conclusion}
In this paper, we have proposed an autonomous disinfection system based on a mobile robot manipulator, with a UV dose optimization strategy that can dynamically adjust the corresponding illuminance according to the level of virus concentration. Ensuring that all objects in the environment receive adequate UV exposure also significantly reduces the time it takes to disinfect. A series of experiments have been conducted on the intelligent ward in our university for performance evaluation. The experimental results have demonstrated its effectiveness and precision for disinfection. In future work, we plan to apply the proposed system for a field study in a hospital to further enhance its practicality.
%%%%%%%%%%%%%%%%%%%%%%%%%%%%%%%%%%%%%%%%%%%%%%%%%%%%%%%%%%%%%%%%%%%%%%%%%%%%%%%%
\section*{Acknowledgment}
This work was financially supported in part by the National Science and Technology Council, Taiwan.

%%%%%%%%%%%%%%%%%%%%%%%%%%%%%%%%%%%%%%%%%%%%%%%%%%%%%%%%%%%%%%%%%%%%%%%%%%%%%%%%

\end{document}